\pdfoutput=1

\documentclass[11pt]{article}

\usepackage{authblk}
\usepackage[review]{acl}

\usepackage{times}
\usepackage{latexsym}
\usepackage[T1]{fontenc}

\usepackage[utf8]{inputenc}

\usepackage{microtype}

\usepackage{graphicx}
\usepackage{booktabs}
\usepackage{amsfonts,amssymb}
\usepackage{multirow}
\usepackage{bm}
\usepackage{stfloats}

\title{Fine-grained Contrastive Learning for Definition Generation}
\renewcommand\footnotemark{}
\author{\textbf{Hengyuan Zhang\textsuperscript{1*}\thanks{\textsuperscript{*} Equal contribution}, Dawei Li\textsuperscript{2*}, Shiping Yang\textsuperscript{3}, Yanran Li\textsuperscript{4\dag}}\thanks{\textsuperscript{\dag} Corresponding author} \\
  \textsuperscript{1}Shenzhen International Graduate School, Tsinghua University \\
  \textsuperscript{2}Halicioğlu Data Science Institute, University of California, San Diego \\
  \textsuperscript{3}School of Computer Science, Beijing University of Posts and Telecommunications\\ 
  \textsuperscript{4}Independent Researcher \\
  \texttt{zhang-hy22@mails.tsinghua.edu.cn,} \quad \texttt{dal034@ucsd.edu,} \\
  \texttt{yangshiping@bupt.edu.cn,} \quad \texttt{yanranli.summer@gmail.com} \\
}

\begin{document}
\maketitle
\begin{abstract}
 Recently, pre-trained transformer-based models have achieved great success in the task of definition generation (DG). However, previous encoder-decoder models lack effective representation learning to contain full semantic components of the given word, which leads to generating under-specific definitions. To address this problem, we propose a novel contrastive learning method, encouraging the model to capture more detailed semantic representations from the definition sequence encoding. According to both automatic and manual evaluation, the experimental results on three mainstream benchmarks demonstrate that the proposed method could generate more specific and high-quality definitions compared with several state-of-the-art models.
\end{abstract}

\section{Introduction} \label{introduction}
When readers find some expressions unfamiliar during reading a text, machines can help. The task of Definition Generation (DG) aims to generate a textual definition for a given word or phrase (the target), according to a surrounding context (the local context)~\cite{ni-wang-2017-learning}. In addition to assisting readers in comprehending expressions, the task of DG is also useful for generating definition when building dictionaries. 


Recently, pre-trained encoder-decoder models have achieved great successes on this task~\cite{huang2021definition,kong2022multitasking}. Despite their successes, the definitions produced by these pre-trained models often contain several types of errors~\cite{noraset2017definition,huang2021definition}. According to Table~\ref{pic:error ratio}, ``under-specific problem'' is the most frequent error that the generated definition conforms to the general semantics but loses certain parts of meaning of the target word. As presented in Table~\ref{under-specific case}, the definition produced by T5 model is under-specific as it omits the meaning of \emph{great} in the word ``double'' under the context ``\emph{ate a double portion}''. The under-specific problem harms the accuracy of the generated definitions and in turn limits the applications of definition generation techniques in many scenarios. 

\begin{table}[t]
\centering
\setlength{\tabcolsep}{11mm}{
\begin{tabular}{@{}cc@{}}
\toprule
Error Types          & Ratio \\ \midrule
\bf{Under-spcified}       & \bf{9.0}\% \\
Over-specified       & 5.5\% \\
Self-reference       & 3.0\% \\
Wrong part-of-speech & 1.0\% \\
Opposite             & 1.0\% \\ \bottomrule
\end{tabular}}
\caption{Ratio of each error type of the definitions generated in \citet{huang2021definition}.}
\label{pic:error ratio}
\end{table}


This problem is partially attributed to the decoder's inability to fully extract the semantic components from the word encoding~\cite{li2020explicit}. For pre-trained encoder-decoder models, they focus on restoring and denoising the whole text in the pre-training stage, rather than learning fine-grained semantic representation of a single word or phrase~\cite{lewis2019bart,bi2020palm,shao2021cpt}. In other words, the pre-trained encoder-decoder models are ineffective in capturing rich semantic components for the given word thus leading to generating under-specific definitions.


\begin{table}[!t]
\centering
\begin{tabular}{l|l}
\hline
\emph{word}      & double                                                                                   \\ \hline
\emph{Reference} & twice as great or many                                                                   \\ \hline
\emph{Generated} & \begin{tabular}[c]{@{}l@{}}characterized by two equal parts\\ or components\end{tabular} \\ \hline
\end{tabular}
\caption{The definition of word ``double'', where \emph{Reference} is from WordNet dictionary and \emph{Generated} is by T5-Base of \citet{huang2021definition}.\label{under-specific case}
}
\end{table}

To remedy the under-specific problem in pre-trained definition generation models, we get inspired from contrastive learning method~\cite{radford2021learning,li2020oscar} and propose a novel definition generation method based on a designed contrastive objective. Conceptually, definition generation is to transform the encoding of the target word to its textual interpretation. To this end, the encoding and the decoding of the target word can be regarded as two views of representations with respect to the same semantics. Our idea is then to leverage the two representations in the definition generation model, and encourage them to align with each other to capture fine-grained semantics. Specifically, we treat the target word representation and the definition representation as a positive pair, and feed them into a contrastive learning objective. This kind of contrastive loss is naturally complementary for the language generation loss, and can be seamlessly incorporated into existing pre-trained encoder-decoder models. 

To validate the effectiveness of our proposal, we conduct a series of experiments on three publicly available datasets. Both automatic and manual evaluation results suggest that our method generates more specific definitions and addresses well the under-specific problem in the task of definition generation. In general, our contributions can be summarized as follows:

\begin{itemize}
\item We tackle the under-specific problem for pre-trained definition generation models by developing a novel fine-grained contrastive learning objective. 

\item We validate the effectiveness of the proposed method through comparing with several SOTA models on three popular datasets using both automatic and manual judgments.\footnote{Our code could be found in \url{https://github.com/rattlesnakey/Definition-Gneration-Contrastive}}

\item We analyze the details of our method by performing ablated studies and demonstrate the effect of our method in addressing under-specific problem based on case studies.

\end{itemize}

\section{Related Work}

\subsection{Definition Generation}
The task of Definition Generation is firstly proposed by~\citet{noraset2017definition}. They used word embedding to generate its corresponding definition, and utilize definition generation as an auxiliary task for reverse dictionary and word embedding training. Some later works explore more application scenarios and model architectures for definition generation. \citet{ni-wang-2017-learning} propose a dual-encoder model to generate the proper definition of the given word under a specific context, and use it for explaining emerging words on the Internet. \citet{gadetsky2018conditional} use both local and global information of the words in their model for word disambiguation. Following them, \citet{ishiwatari2019learning} design gate mechanisms to fuse multi-source information of the word and context. Furthermore, some works attempt to utilize other information of the target word. \citet{washio2019bridging} build relation of defined and defining words using word pair embedding \cite{joshi2018pair2vec}. Different from former works that using distributed representations of target words, \citet{yang2019incorporating} introduce target words' concepts in HowNet~\cite{dong2003hownet} as fine-grained knowledge in Chinese definition modeling. Also, there exist literature works based on refined methods to learn the target words. Both \citet{li2020explicit} and \citet{reid2020vcdm} decompose the meaning of the target word into a group of latent variables and rely on variational inference for estimation. 

Recently, pre-trained encoder-decoder models have been used in definition generation and achieved great success. \citet{bevilacqua2020generationary} use special tokens to mark the target word in the context and feed them into a BART model~\cite{lewis2019bart}. \citet{huang2021definition} fine-tune a T5 model and re-rank all the candidate results from the T5 model to obtain definitions in a proper specificity. \citet{kong2022multitasking} design a MASS model based on multi-task framework to generate simple definition in an unsupervised manner. Despite of their promising performances on definition generation, the under-specific problem has been less investigated. Although \citet{huang2021definition} design a scoring mechanism that measures definitions' specificity, we argue that the fundamental reason of the under-specific problem lies in the lack of fine-grained semantic learning in pre-trained encoder-decoder models, which we leverage contrastive learning to address in this work.

\subsection{Contrastive Learning in Semantic Representation}
Contrastive learning has been widely used in enhancing semantic information for various NLP tasks. 
For example, \citet{gao2021simcse} use a dropout trick to derive positive samples in the embedding level, and then apply both supervised and self-supervised methods to acquire better sentence embedding.
\citet{radford2021learning} use contrastive learning to pre-train a vision language model to align the message between images and their corresponding text.
\citet{li-etal-2022-lingjing} use masked language modeling and contrastive learning to perform multi-task pre-training, and demonstrate that contrastive learning benefits in connecting word gloss and its corresponding vectors.
\citet{li2020oscar} and \citet{srivastava-vemulapati-2022-tldr} implement contrastive learning as an auxiliary task to encourage the transformer encoder better capture the semantic alignment.


In this work, we borrow the idea of using contrastive methods in semantic representation learning. For a given target word, there are two representations in the task of definition generation: the word representation generated by the encoder, and the definition representation produced by the decoder. These two kinds of representations can be regarded as two views of the semantics of the target word to be explained. By aligning the representation spaces between the encoder and the decoder using contrastive learning, we force the model to pay much attention to the fine-grained semantic information during representation learning. In this way, the under-specific problem will be mitigated when using pre-trained encoder-decoder models to generate definitions.

\section{Method}
\label{Method}

\begin{figure*}[!h]
    \centering
    \includegraphics[width=16cm]{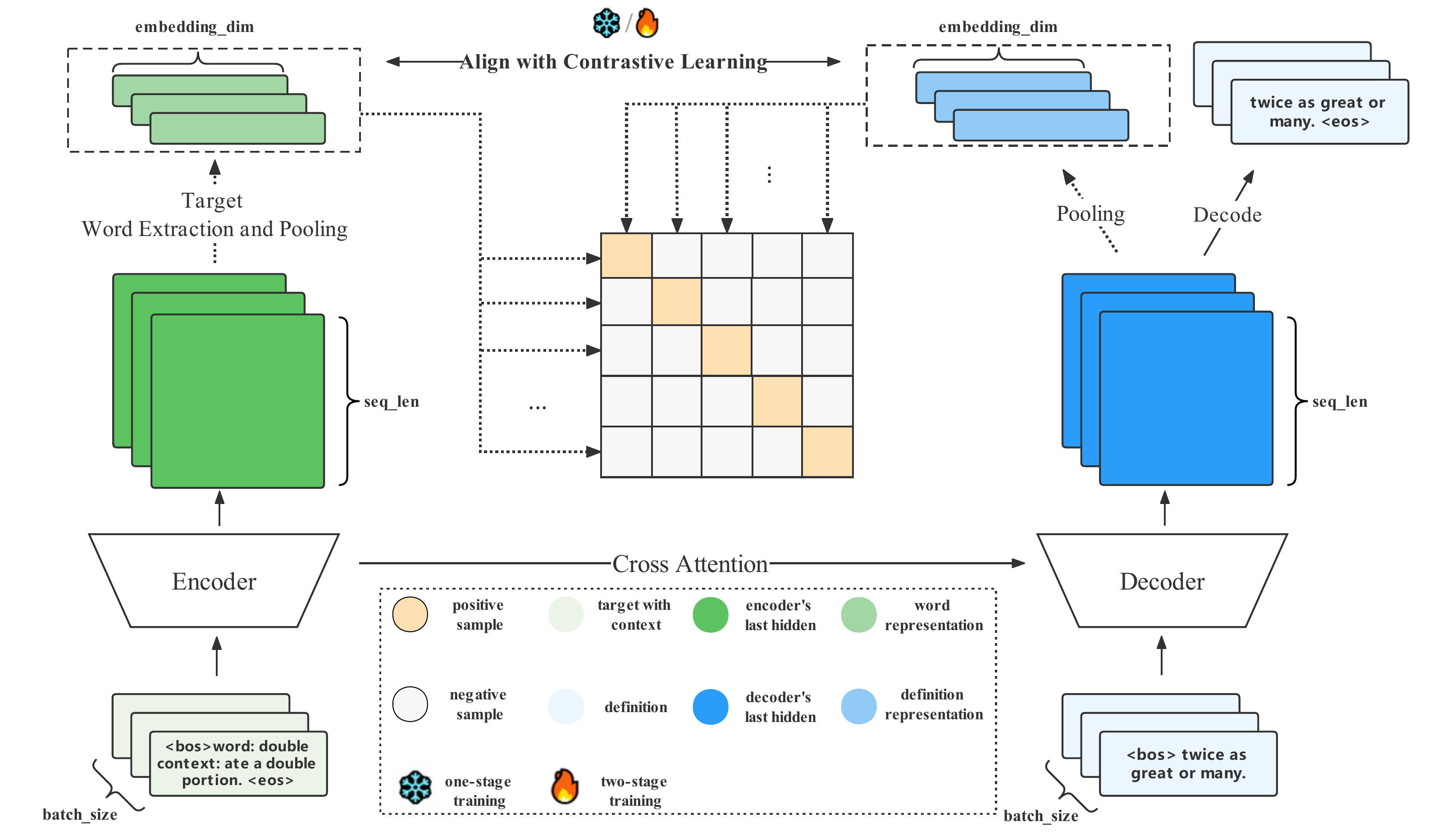}
    \caption{The overview training process of our proposed model. The solid arrows indicate the data-flow of maximum likelihood estimate learning, and the dash arrows indicate the data-flow of contrastive learning. Note that the snow icon represents the one-stage training where the model is trained from scratch with the contrastive and generation loss. The fire icon represents the two-stage training, where at the first stage the model is fine-tuned only by the generation loss, and at the second stage the contrastive and generation loss are together applied then.}
    \label{overview}
\end{figure*}

In this section, we present our method of using contrastive learning to enhance target words' representation for definition generation. Specifically, we first formulate the definition generation task and introduce the denotations (Section~\ref{Task Formulation}). Then we provide a preliminary description of the definition generation processing based on T5 (Section~\ref{Definition Generation with T5}). Finally, we explain how to apply the contrastive loss in the training process to solve the under-specific problem and improve the generation quality (Section~\ref{Contrastive Learning}). Figure~\ref{overview} depicts the overview pipeline of our method.

\subsection{Task Formulation}
\label{Task Formulation}
Given a word or phrase $W = \{w_i,...,w_j\}$ and its surrounding context $C = \{w_0,...,W_k\}(0<i<j<k)$, the task of definition generation is to generate the definition $D = \{d_0,...d_T\}$ to explain the meaning of $W$ under $C$. This process can be formulated as:
\begin{equation}\label{task formulation}
P(D|W,C) = \prod \limits_{t=0}^T p(d_t|d_{<t},W,C)
\end{equation}

\subsection{Definition Generation with T5}
\label{Definition Generation with T5}
Our work aims at addressing the under-specific problem when using pre-trained encoder-decoder models for definition generation. Without loss of generality, we take T5~\cite{raffel2020exploring} as our backbone model, which is a transformer-based encoder-decoder model trained on large-scale corpus, and has demonstrated its effectiveness on definition generation task~\cite{huang2021definition}.

To apply T5 for definition generation, we first concatenate the target word and the given context together behind the prefix prompts ``word:'' and ``context:'' respectively. The concatenated input is then fed to the T5 encoder with $L_E$ layers of encoder block $\rm E\_Block$. Then we get the last hidden state $\mathbf{H}^{ L_E}$, which contains the semantic information of the target word and local context:


\begin{equation}\label{encoder}
\mathbf{H}_0={\rm Emb}({\rm Splice}(W, C))
\end{equation}

\begin{equation}\label{encoder}
\mathbf{H}^l={\rm E\_Block}(\mathbf{H}^{l-1}),l\in[1,L_E]
\end{equation}

Here $W$ stands for the target word, $C$ for the given context, and $\rm Splice$ is the operation to concatenate the target word and the given context with their corresponding prefixes. Also, $\rm Emb$ is the Embedding layer that converts the input tokens into embedding vectors.

After encoding, the T5 decoder will learn to generate an appropriate definition conditioned on encoding $\mathbf{H}^{L_E}$ and the previous generation result. During decoding, the teacher-forcing mechanism is applied to guarantee the previous information being attended at the current step $t$:

\begin{equation}\label{decoder}
\mathbf{G}_t^{0}={\rm Emb}(D_t)
\end{equation}

\begin{equation}\label{decoder}
\mathbf{G}_t^{l}={\rm D\_Block}(\mathbf{H}^{L_E},\mathbf{G}_{\leq{t}}^{l-1}),l\in[1,L_D]
\end{equation}

Here $D_t$ represents the $t^{th}$ token in the definition sequence. After passing through $L_D$ layers of the decoder block $\rm D\_Block$, we get the decoder's last hidden state $\mathbf{G}^{L_D}$. 

Finally, a softmax function is added upon a linear head to transform $\mathbf{G}^{L_D}$ into a prediction distribution matrix $\mathbf{V}\in \mathbb{R}^{|V|\times |D|}$. Here $|V|$ and $|D|$ stand for the vocabulary size and the length of the ground-truth definition, respectively. To optimize, a cross-entropy loss is applied to measure the discrepancy between the generated distribution and the ground-truth distribution.

\subsection{Fine-grained Modeling with Contrastive Learning}
\label{Contrastive Learning}
Here we describe our proposal of applying contrastive learning for definition generation. Conceptually, definition generation requires the model to understand the target word and produce its definition to explain the meaning of the word in the context. To this end, definition generation can be cast as a mapping between the understanding of the target word (in the encoder side) and the generation of the word definition (in the decoder side).

Hence, our idea is to leverage the representations obtained from both the encoder and decoder side in the model, and encourage them to align with each other to capture fine-grained semantics. By regarding the representations from both sides as two views of the target words' semantics, we are able to deploy a contrastive loss during the generation process in the training phase.

Formally, we denote the {target word encoding} generated by the encoder in T5 as $\mathbf{H}_{target}$, and the {definition encoding} generated by the decoder as $\mathbf{G}^{L_D}$. In general, target word encoding $\mathbf{H}_{target}$ is obtained by extracting the encoding of the target word's position in $\mathbf{H}^{L_E}$, and
definition encoding $\mathbf{G}^{L_D}$ is generated by the decoder to decode and get the definition sequence later. 

\begin{table*}[!t]
\centering
\begin{tabular}{@{}lccccccccc@{}}
\toprule
               & \multicolumn{3}{c}{WordNet}                                                      & \multicolumn{3}{c}{Oxford}                                                       & \multicolumn{3}{c}{Urban}                                                        \\ \midrule
\textbf{}      & \multicolumn{1}{l}{Train} & \multicolumn{1}{l}{Valid} & \multicolumn{1}{l}{Test} & \multicolumn{1}{l}{Train} & \multicolumn{1}{l}{Valid} & \multicolumn{1}{l}{Test} & \multicolumn{1}{l}{Train} & \multicolumn{1}{l}{Valid} & \multicolumn{1}{l}{Test} \\ \midrule
Phrases        & 7,938                     & 998                       & 1,001                    & 33,128                    & 8,867                     & 8,850                    & 190,696                   & 26,876                    & 25,797                   \\
Entries        & 13,883                    & 1,752                     & 1,775                    & 97,855                    & 12,232                    & 12,232                   & 411,384                   & 57,883                    & 36,450                   \\
Context length & 5.81                      & 5.64                      & 5.77                     & 17.74                     & 17.80                     & 17.56                    & 10.89                     & 10.86                     & 11.22                    \\
Desc. length   & 6.61                      & 6.61                      & 6.85                     & 11.02                     & 10.99                     & 10.95                    & 10.99                     & 10.95                     & 12.05                    \\ \bottomrule
\end{tabular}
\caption{Statistics of The Datasets.}
\label{Statistics of datasets}
\end{table*}

After encoding, we use a pooling function $\rm f()$ to aggregate the $\mathbf{H}_{target}$ and $\mathbf{G}^{L_D}$ respectively, and obtain {target word representation} $\mathbf{h}$ and {definition representation} $\mathbf{g}$ with the same length:

\begin{equation}\label{}
\mathbf{h} = {\rm f}(\mathbf{H}_{target})
\end{equation}

\begin{equation}\label{}
\mathbf{g} = {\rm f}(\mathbf{G}^{L_D})
\end{equation}

Note that there are multiple choices to implement the pooling function $\rm f()$. Empirically motivated, we adopt max-pooling $\rm Max()$ and achieve our best performance in the main experimental results. We also present the results with mean-pooling $\rm Mean()$ in the ablation study in the following sections. 

Eventually, we treat the two representations $\mathbf{h}$ and $\mathbf{g}$ in the same sample as a positive pair, and define our contrastive learning stage's training objective as follows:

\begin{equation}\label{eq:contrastive}
L_C = \sum_{i=1}^N -log\frac{e^{{\rm sim}(h_i,g_i)}/\tau}{\sum_{j=1}^N e^{{\rm sim}(h_i,g_j)}/\tau}
\end{equation}
where $N$ denotes a mini-batch of training samples. The $\tau$ is a temperature hyper-parameter and $\rm sim(,)$ stands for the cosine similarity function. During learning, the contrastive loss in Eq.~\ref{eq:contrastive} enforces the model to concentrate on the discrepancy between the two views of the same semantic unit, i.e., the target word.

\subsection{Two-Stage Training}
\label{Two-Stage Training}
In addition to the newly introduced contrastive loss, we also train the model based on the commonly adopted generation loss, which takes advantage of language modeling ability.

As depicted in Figure~\ref{overview}, our full training strategy follows a two-stage paradigm. At the first stage, we finetune our model only with the generation loss. In the second stage, we combine the contrastive loss in the training and optimize the model with mixed loss $L_{Final}$:

\begin{equation}\label{eq:Final_Loss}
L_{Final} = \lambda*L_C + (1-\lambda)*L_G
\end{equation}
where $\lambda$ is a hyper-parameter to balance the two loss. The two-stage training allows to incrementally train the decoder learn the semantic information from the definition sequence at the very beginning, and guarantees the quality of the definition encoding for the encoder to discriminate in the following stage. 
 
By combining the contrastive loss with generation loss, our method is able to: (1) learn fine-grained representation for the target word, (2) mitigate the under-specific problem in the encoder-decoder models, and (3) improve the overall quality of the generated definition.

\section{Experiments}
In this section, we compare our method with several state-of-the-art methods and conduct a series of experiments to verify the effectiveness of our method in addressing the under-specific problem in definition generation.

\subsection{Datasets}
\label{Datasets}

For evaluation, we follow previous works and acquire three popular datasets, which are ensembled by \citet{ishiwatari2019learning}\footnote{\url{http://www.tkl.iis.u-tokyo.ac.jp/~ishiwatari/naacl_data.zip}}.
Each entry in a dataset consists of three elements: (1) a target word or phrase, (2) the corresponding definition, and (3) one usage example of the target as a local context. If a target has multiple definitions and examples, we treat them as different entries. For fair comparison, each dataset is split into $\textit{train}$, $\textit{dev}$ and $\textit{test}$ sets according to \citet{ishiwatari2019learning}. The statistics of these datasets are shown in Table~\ref{Statistics of datasets}.

\paragraph{WordNet dataset} The Wordnet dataset is collected by \citet{noraset2017definition} from the Wordnet dictionary and the GNU Collaborative International Dictionary of English\footnote{\url{http://wwwgcide.gnu.org.ua}}. In this work, we follow \citet{ishiwatari2019learning} and use the extended version of WordNet dataset, where usage examples for each entry are added and the entries without usage examples are removed.

\paragraph{Oxford dataset}
The Oxford dataset is collected using APIs of Oxford Dictionaries\footnote{\url{https://developer.oxforddictionaries.com}} by \citet{gadetsky2018conditional}.

\paragraph{Urban dataset}
The Urban dataset is collected from Urban Dictionary\footnote{\url{https://www.urbandictionary.com}}, which is the largest online slang dictionary. Unlike the former two datasets, this dataset contains many non-standard phrases with more than one word. In Urban dataset, all terms, definitions, and examples are submitted by users on the Internet. 

\subsection{Compared Models}

To evaluate the effectiveness of our method, we compare with the following models:

\noindent\textbf{Global} \cite{noraset2017definition} is the first definition generation technique that only accesses the global context of the target word.

\noindent\textbf{Local} \cite{ni-wang-2017-learning} is the refined model that utilizes both word-level and character-level information to get the target word encoding based on the surrounding context.

\noindent\textbf{I-Attention} \cite{gadetsky2018conditional} combines local and global contexts together and employs latent variable modeling and soft attention mechanisms.

\noindent\textbf{LOG-CaD} \cite{ishiwatari2019learning} integrates the designs in the previous methods and uses gate-mechanism to balance information from different sources in the decoding phase.

\noindent\textbf{T5-Reranking} \cite{huang2021definition} is the current SOTA method in definition generation. It uses a pre-trained T5 to get generation results first and designs a score mechanism to measure and sample definitions in appropriate specificity. 

\noindent\textbf{T5-Base}
Besides, we also fine-tune a pre-trained T5 only using the generation loss we mention in Section \ref{Two-Stage Training} as a baseline (denoted as T5-Base).


\begin{table*}[!t]\small
\centering
\setlength{\tabcolsep}{2.5mm}{
\begin{tabular}{@{}ccccccc@{}}
\toprule
                                    & \multicolumn{2}{c}{WordNet} & \multicolumn{2}{c}{Oxford} & \multicolumn{2}{c}{Urban} \\ \midrule
                                    & BLEU         & NIST         & BLEU         & NIST        & BLEU        & NIST        \\
\toprule
I-Attention                          & 23.77        & 44.30        & 17.45        & 35.79       & 8.81        & 19.43       \\
Local                          & 24.78        & 40.32        & 17.58        & 31.30       & 8.99        & 17.39       \\
Global                         & 23.59        & 49.70        & 14.95        & 32.79       & 5.15        & 10.45       \\
LOG-CaD                           & 25.19        & 43.54        & 18.57        & 38.22       & 9.93        & 19.29       \\
T5-Reranking      & \bf{32.72}        & 64.57        & 26.52        & 74.17       & 17.71       & 35.53       \\ \bottomrule
T5-Contrast (Ours) &     32.05$_{(-2.1\%)}$         &       \bf{74.71}$_{(+15.5\%)}$       & \bf{27.11}$_{(+2.2\%)}$        & \bf{79.42}$_{(+7.1\%)}$       & \bf{19.44}$_{(+9.8\%)}$       & \bf{41.01}$_{(+15.4\%)}$       \\
T5-Base &     31.72         &      57.35        & 25.44        & 66.92       & 17.66       & 26.86       
\\\bottomrule
\end{tabular}}
\caption{Automatic evaluation results on test sets of three datasets. The best results in each dataset are in bold. We also add the quantitative comparison results between our method and the strongest baseline model T5-Reranking.}
\label{Experimental_Results}
\end{table*}



\subsection{Automatic Metrics}
Following common practice, we adopt two automatic evaluation metrics to assess the quality of the definitions generated by each model.
\paragraph{BLEU}
The metric BLEU \cite{papineni2002bleu} has been widely used in previous works to measure the closeness between the generated results and human reference. It measures the geometric average of the precision over hypothesis n-grams with an additional penalty to discourage short definition.
\paragraph{NIST}
NIST~\cite{doddington2002automatic} is similar to BLEU, but considers up-weighting rare, informative n-grams. We use NLTK\footnote{\url{https://www.nltk.org}} tool to calculate NIST metric.

\subsection{Experimental Setups}
\label{Settings}
We train all models in PyTorch\footnote{\url{https://github.com/pytorch/pytorch}} \cite{paszke2019pytorch}, and use the HuggingFace\footnote{\url{https://github.com/huggingface/}} \cite{wolf2019huggingface} implementation of T5. We train each model on a V100 GPU. For compared models, we replicate experiments following the implementations details released by \citet{huang2021definition}. For training our model, we use the base version of T5 with the same size of \citet{huang2021definition}. 
For each dataset, we finetune it using Adam ~\cite{kingma2014adam} optimizer with an initial learning rate of 3e-4 and the batch size of 16. In all the experiments, we train our model with a two-stage strategy as described in the previous section. Please refer to Appendix~\ref{Detailed Training Setting} for the detailed training settings in each stage, like max-epoch and early-stop threshold.




\subsection{Main Results}
\label{Experimental Results}


Table~\ref{Experimental_Results} shows the automatic comparison results of each compared model on the three datasets. Considering the absolute scores, the proposed method T5-Contrast significantly outperforms other 5 models on almost every metric across the three datasets. Although the BLEU score on WordNet dataset obtained by our method is slightly lower (2.09\%) than T5-Reranking \cite{huang2021definition}, the NIST score of our method in WordNet dataset is notably higher (15.70\%) than theirs. This strongly demonstrates the effectiveness and generalization of the proposed method in generating high-quality definitions for a given word under a context. 

It is obvious that the two refined T5 model, e.g., T5-Reranking~\cite{huang2021definition} and T5-Contrast (ours) are the best and the second best model. By comparing the relative increases between these two models, we notice that our method T5-Contrast improves a lot on Urban dataset (9.8\% relative increase on BLEU, and 15.4\% relative increase on NIST). As compared to the datasets WordNet and Oxford, Urban dataset is more challenging due to the targets in it are often phrases, and the definitions are often long and complex. Drawing on the great promotion by T5-Contrast (ours) on the difficult dataset, we highlight the necessity of modeling fine-grained semantic in pre-trained models for definition generation.

\subsection{Ablation Study}

\begin{table*}[!t]
\centering
\label{tab-Experimental Results}
\setlength{\tabcolsep}{5.5mm}{
\begin{tabular}{@{}ccccccc@{}}
\toprule
                                    & \multicolumn{2}{c}{WordNet} & \multicolumn{2}{c}{Oxford} & \multicolumn{2}{c}{Urban} \\ \midrule
                                    & BLEU         & NIST         & BLEU         & NIST        & BLEU        & NIST        \\
\toprule
Ours                          & \bf{32.05}        & \bf{74.71}      & 27.10        & 79.42        & \bf{19.44}        & \bf{41.01}              \\
w/ $\rm Mean()$                          & 31.07        & 71.48      & \bf{27.13}        & \bf{80.33}        & 18.57        & 40.29             \\
w/ One-stage training                         & 31.75        & 73.79      & 27.06        & 79.90        &  16.49       &   31.46           \\
\bottomrule
T5-Base                           & 31.72        & 57.35        & 25.44        & 66.92        & 17.66        & 26.86              \\
\bottomrule
\end{tabular}}
\caption{Ablation study results on test sets. The best numbers are in bold.}
\label{ablation study}
\end{table*}

As introduced in Section~\ref{Method}, there are two novel designs in our method: (1) the contrastive learning with a pooling function, and (2) a two-stage training strategy that combines both generation loss and the contrastive loss. In this subsection, we conduct an ablation study to examine the variants of each component in the proposed method.

As shown in Table~\ref{ablation study}, replacing the pooling function $Max()$ with the mean-pooling $Mean()$ will bring in different changes on different datasets. Whereas the automatic scores drop a lot on WordNet and Urban datasets, they increase a bit on Oxford dataset. This indicates that the choice of pooling function might be empirically motivated, and in general the effect of contrastive learning does not vary a lot when the pooling function changes.

Moreover, we also examine the importance of two-stage training by removing the first stage of generation-only training and directly training our model using the combined loss (One-stage training).
Especially on the challenging Urban dataset, the performance dramatically decreases when training T5 from scratch using the combined loss. Last but not least, each of our ablated variant still surpasses T5-Base on most metrics, which indicates the method's robustness.

\subsection{Analysis on Hyper-Parameter}

To explore how our method would be affected by the choice of the hyper-parameter $\lambda$ in Eq.~\ref{eq:Final_Loss}, we remain other settings the same as we mentioned in Section~\ref{Settings} and set different $\lambda$ for each model to observe the performance change. The results on the Oxford dataset are reported in Table~\ref{hyper-param}. As shown, when $\lambda$ is set to 0.0, the model is ``degraded'' to the compared T5-Base model. Considering T5-Base model is fine-tuned only using the generation loss in our setting, it is identical to a variant without contrastive loss in the second training stage. To this end, their performances are the same. Also, the performance of the model when $\lambda$ is set to 1.0 (without generation loss in the second training stage) is pretty bad. We attribute it to the fact that our task requires the ability of language generation and thus still need generation loss to guide contrastive learning in the right way. Besides the above extreme values of $\lambda$, we find the model achieves better performance when $\lambda$ is higher ($\lambda$=0.8 and $\lambda$=0.6). It further illustrates that after the first stage of generation-only training, the model will benefit more from our fine-grained contrastive learning.

\begin{figure}[htbp]
    \centering
    \includegraphics[width=7.6cm]{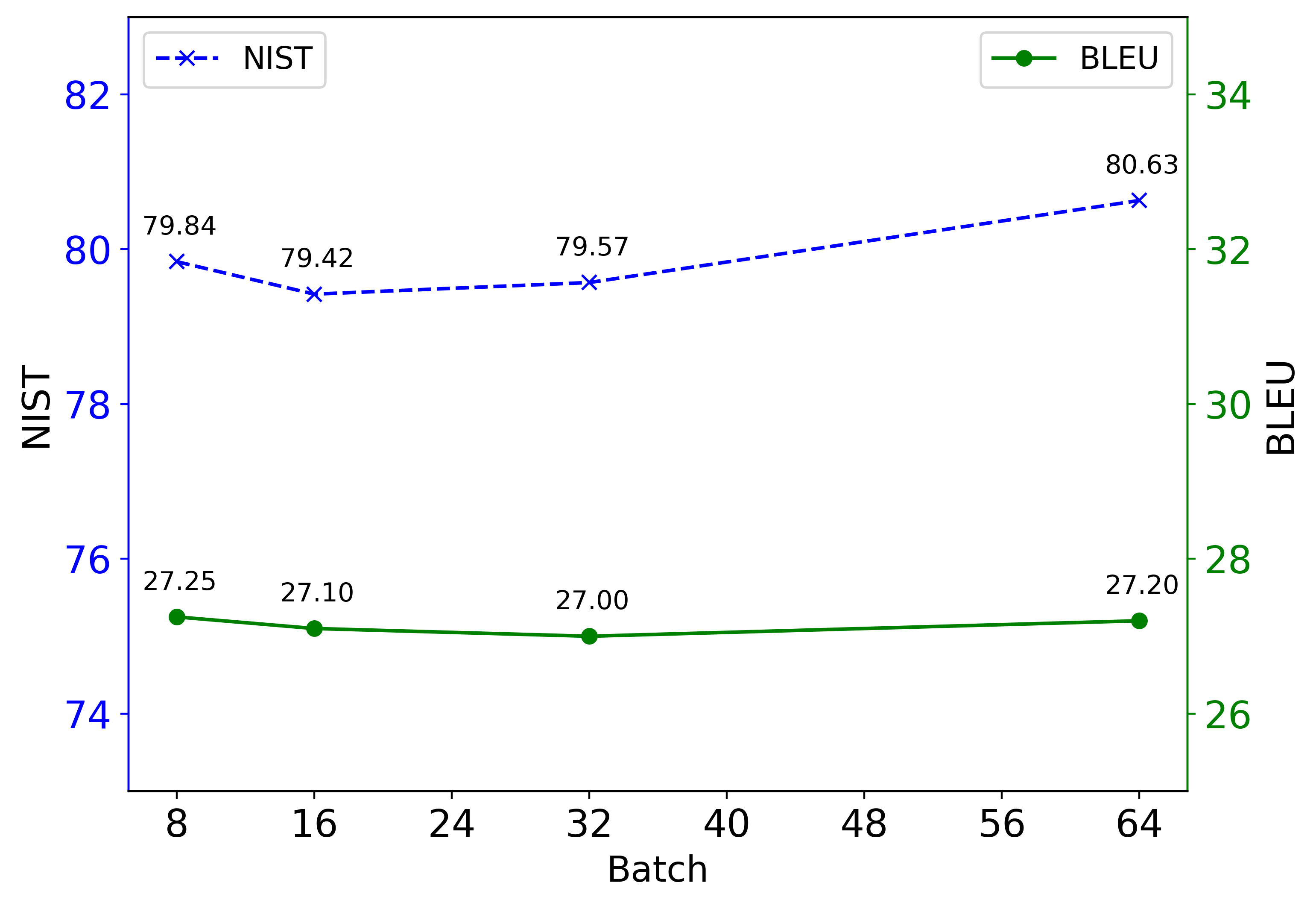}
    \caption{Training batch size analysis results on Oxford test set.}
    \label{fig:batch size}
\end{figure}

We also investigate the influence of training batch size on our method. We set our training batch size $\in \{8, 16, 32, 64\}$ and conduct experiments on the Oxford dataset. As Figure \ref{fig:batch size} shows, each model's performance in different batch size settings doesn't show much difference. It is probably due to our proposed method base on pre-trained T5 which has good prior knowledge. 


\begin{table}[htbp]
\centering
\begin{tabular}{c|c|c}
\hline
$\lambda$   & BLEU & NIST \\ \hline
1.0 & 7.71 & 25.67 \\ \hline
0.8 &  27.11    &   79.42   \\ \hline
0.6 &   27.23   &    79.86  \\ \hline
0.4 &   26.66   &    77.68  \\ \hline
0.2 &    26.54  &    78.60  \\ \hline
0.0 & 25.44 & 66.92 \\ \hline
\end{tabular}
\caption{Different $\lambda$ settings on Oxford test set.}\label{hyper-param}

\end{table}

\subsection{Manual Evaluation}
To adequately evaluate the generated definitions, we also adopt three kinds of manual metrics: (1) Acccuracy (Acc.) ~\cite{ishiwatari2019learning} learning measures the semantic similarity between the generated definitions and the target words; (2) Fluency (Flu.) evaluates the readability of the generated definitions without considering the semantic alignment; (3) Under-specified (Under-spec.)~\cite{noraset2017definition}  calculates the ratio of under-specific definitions in the generated cases, which is curated to assess the model's capabilities in addressing the under-specific problem. The lower the ratio is, the better the model is in capturing fine-grained semantics during definition generation. Note that both Acc. and Flu. metrics are likert-scale of \{0,1,2,3,4,5\}.

Considering the labor resource cost, we conduct manual evaluation on Oxford dataset, and only compare our method with the strongest baseline T5-Reranking and the backbone T5-Base. For fairness, we randomly select 100 samples, acquire the generation results of each compared model, and pair them with the Golden definitions. Then we ask three well-trained annotators with at least CET (College English Test) 6-level English skills to rate the generated definitions according to the three manual metrics. At last, each model's score is the average of the three annotators' rates and the agreement among the annotators is ICC 0.962 with (p<0.001)~\cite{bartko1966intraclass}, which indicates the results are reliable enough.

According to Table~\ref{manual-evaluation}, the definitions generated by the proposed method T5-Contrast are better than those by other two models in terms of all the three metrics. Notably, the under-specific ratio significantly drops from 7.6\% (T5-Base) to 4.8\% (Ours). The manual evaluation results imply that the definitions produced by our method are more accurate, fluent, and fine-grained as compared to other pre-trained models.

\begin{table}[htbp]
\centering
\begin{tabular}{@{}cccc@{}}
\toprule
        & Acc.  & Flu.  & Under-spec. \\ \midrule
T5-Base & 3.17 & 3.89 & 7.6\%          \\
T5-Reranking & 3.43 & 3.95 & 5.4\%          \\
T5-Constrast (Ours)    & 3.46 & 4.03 & 4.6\%          \\
Golden  & 4.57 & 4.92 & 0.2\%          \\ \bottomrule
\end{tabular}
\caption{Manual evaluation results on Oxford dataset.}\label{manual-evaluation}
\end{table}

\subsection{Case Study}

For better understanding, we show some example definitions generated by these compared models in Table~\ref{tab:case study}. It is obvious that T5-Base produces an under-specific definition ``\emph{a positive criticism}'' for the target word ``\emph{praise}'' in the context \emph{he always appreciated praise for this work}. The generated definition roughly expresses the positive meaning of the target word \emph{appreciate}, but fails to provide the accurate meaning of \emph{approval and commendation} in \emph{praise}. In this case, this example definition by T5-Base is under-specific. As for T5-Reranking, it generates the word ``\emph{goodwill}'', which is a multi-sense word where the one sense is ``a kindly feeling of support'' and the other sense is ``the favor or advantage of a business''. As such, this definition by T5-Reranking is also inaccurate to describe the word \emph{praise}. On the contrary, the definition generated by our model that ``\emph{an expression of admiration or approval}'' is more specific, which shows the effectiveness of our proposed method to remedy the under-specific problem. Due to the space limit, we give more sampled examples in Appendix \ref{Additonal Case Study}. 

It is also worth noting that, with our contrastive learning loss, some definitions generated in the test time are even identical with their ground truths. This also supports our idea that fine-grained contrastive learning will benefit the pre-trained encoder-decoder models in modeling and generating definitions. We also put these kinds of cases in Appendix \ref{Perfectly Reproduced Examples}.


\begin{table}[htbp]
\centering
\begin{tabular}{c|l}
\hline
\emph{Word}         & Praise                                                                                 \\ \hline
\emph{Context}      & \begin{tabular}[c]{@{}l@{}}He always appreciated \bf{praise} \\ for his work.\end{tabular}  \\ \hline
\emph{T5-Base}      & A \textcolor[RGB]{244,67,54}{positive} critisism.                                                                  \\ \hline
\emph{T5-Reranking} & An act to \textcolor[RGB]{76,175,80}{express} \textcolor[RGB]{244,67,54}{goodwill}.                                                            \\ \hline
\emph{Ours}         & \begin{tabular}[c]{@{}l@{}}An \textcolor[RGB]{76,175,80}{expression} of admiration \\ or \textcolor[RGB]{76,175,80}{approval}.\end{tabular}    \\ \hline
\emph{Ground Truth} & \begin{tabular}[c]{@{}l@{}}An \textcolor[RGB]{76,175,80}{expression} of \textcolor[RGB]{76,175,80}{approval} \\ and \textcolor[RGB]{76,175,80}{commendation}.\end{tabular} \\ \hline
\end{tabular}
\caption{An example showing the two generated definitions for the word ``praise'' by our model, T5-Base and T5-Reranking. The \textcolor[RGB]{76,175,80}{green} text represents the appropriate specificity of the generated definition, and the text in \textcolor[RGB]{244,67,54}{red} represents the hints where the generated definition is under-specific.}
\label{tab:case study}

\end{table}


\section{Conclusion}

In this work, we tackle the under-specific problem when using pre-trained encoder-decoder models for definition generation. To address, We propose a fine-grained contrastive method to inject detailed semantic information into the model. Through extensive experiments, we demonstrate the effectiveness and generalization of the proposed method using both automatic and manual evaluations on three datasets. 

In the future, we aim to introduce more fine-grained methods and language resources into definition generation.

\bibliography{anthology,custom}

\appendix
\onecolumn
\section{Detailed Training Settings}
\label{Detailed Training Setting}


\begin{table*}[h!]
\centering
\begin{tabular}{|c|c|c|c|c|c|}
\hline
\multicolumn{1}{|l|}{Stage} & \multicolumn{1}{l|}{Dataset} & \multicolumn{1}{l|}{Max-epoch} & \multicolumn{1}{l|}{Early-stop} & \multicolumn{1}{l|}{Pooling method} & \multicolumn{1}{l|}{$\lambda$} \\ \hline
\multirow{3}{*}{1}          & WordNet                      & 140                            & 40                              & None                                & 0.0                         \\ \cline{2-6} 
                            & Oxford                       & 50                             & 10                              & None                                & 0.0                         \\ \cline{2-6} 
                            & Urban                        & 30                             & 5                               & None                                & 0.0                         \\ \hline
\multirow{3}{*}{2}          & WordNet                      & 70                             & 40                              & Max                                 & 0.8                         \\ \cline{2-6} 
                            & Oxford                       & 50                             & 10                              & Max                                 & 0.8                         \\ \cline{2-6} 
                            & Urban                        & 15                             & 5                               & Max                                 & 0.8                         \\ \hline
\end{tabular}
\caption{Detailed settings on each of our training stages, including max-epoch, early-stop threshold, pooling method and loss weight $\lambda$.}\label{tab-settings}
\end{table*}

\section{Additional Case Study}
\label{Additonal Case Study}

\begin{table*}[!h]
\centering
\begin{tabular}{@{}ll@{}}
\toprule
\emph{Word}      & underestimate                                                                                                                                            \\
\emph{Context}   & I wish people wouldn't underestimate me, or my strength, or my weakness.                                                                                 \\
\emph{Reference} & regard ( someone ) as less capable than they really are                                                                                                                                                                                              \\
\emph{T5-Base}   & make too low an estimate of                                                                                                                              \\
\emph{Ours}      & make ( someone or something ) appear less important than they really are                                                                                                                                                                             \\ \midrule
\emph{Word}      & line                                                                                                                                                     \\
\emph{Context}   & they gave me a direct line, which was a relief, instead of those infuriating 0800 numbers                                                        \\
\emph{Reference} & a telephone connection or service                                                                                                                        \\
\emph{T5-Base}   & a direct route                                                                                                                                           \\
\emph{Ours}      & a connectionsingle item of telephone service                                                                                                                       \\ \midrule
\emph{Word}      & caution                                                                                                                                                  \\
\emph{Context}   & a man of caution                                                                                                                                         \\
\emph{Reference} & the trait of being cautious                                                                                                                              \\
\emph{T5-Base}   & the trait of being careful                                                                                                                               \\
\emph{Ours}      & the trait of being attentive to possible danger                                                                                                          \\ \midrule
\emph{Word}      & configuration                                                                                                                                            \\
\emph{Context}   & the outcome depends on the configuration of influences at the time                                                                                       \\
\emph{Reference} & an arrangement of parts or elements                                                                                                                      \\
\emph{T5-Base}   & the way in which something is arranged                                                                                                                   \\
\emph{Ours}      & the arrangement of things or events in a system                                                                                                          \\ \midrule
\emph{Word}      & exercise                                                                                                                                                 \\
\emph{Context}   & the doctor recommended regular exercise                                                                                                                  \\
\emph{Reference} & the activity of exerting your muscles in various ways to keep fit                                                                                        \\
\emph{T5-Base}   & the act of working out                                                                                                                                   \\
\emph{Ours}      & the act of participating in regular physical activities                                                                                                  \\ 
\bottomrule
\end{tabular}
\caption{Additional generated cases that showing the effectiveness of our method in solving the under-speciﬁc problem in deﬁnition generation.}
\end{table*}

\onecolumn
\section{Perfectly Reproduced Examples}
\label{Perfectly Reproduced Examples}

\begin{table*}[!h]
\centering
\begin{tabular}{@{}ll@{}}
\toprule
\emph{Word}    & net                                                                                 \\ 
\emph{Context} & the net result                                                                      \\
\emph{Ours}    & conclusive in a process or progression                                                       \\ \midrule
\emph{Word}    & mysterious                                                                          \\ 
\emph{Context} & the new insurance policy is written without cryptic or mysterious terms             \\
\emph{Ours}    & of an obscure nature                                                                \\ 
\emph{Word}    & legally                                                                             \\ 
\emph{Context} & he acted legally                                                                    \\
\emph{Ours}    & in a legal manner                                                                   \\ \midrule
\emph{Word}    & state                                                                               \\
\emph{Context} & state your opinion                                                                  \\
\emph{Ours}    & to express in words        
  \\ \midrule
\emph{Word}    & practically                                                                         \\
\emph{Context} & practically orientated institutions such as business schools                        \\
\emph{Ours}    & in a practical manner                                                               \\ \midrule
\emph{Word}    & passionately                                                                        \\
\emph{Context} & she kissed him passionately                                                         \\
\emph{Ours}    & with passion
  \\ \midrule
\emph{Word}    & nonprofessional                                                                     \\
\emph{Context} & the nonprofessional wives of his male colleagues                                    \\
\emph{Ours}    & not professional                                                                    \\ \midrule
\emph{Word}    & buzz                                                                                \\
\emph{Context} & if you need help debugging it, you're more than welcome to give me a buzz tomorrow. \\
\emph{Ours}    & a telephone call                                                                    \\ \midrule
\emph{Word}    & hereafter                                                                           \\
\emph{Context} & do jews believe in the hereafter such as life after death?                          \\
\emph{Ours}    & life after death                                                                    \\ \midrule
\emph{Word}    & bop                                                                                 \\
\emph{Context} & over 1,000 people bopped, jigged, jived and pogoed to some excellent bands.         \\
\emph{Ours}    & dance to pop music                                                                  \\ \midrule
\emph{Word}    & boo bear                                                                            \\
\emph{Context} & I will love my boo bear ramero forever and always 3                                 \\
\emph{Ours}    & pet name                                                                            \\ \midrule
\emph{Word}    & bang bang                                                                           \\
\emph{Context} & hey chris, do you want to bang bang tonight or will you get marcia'd?               \\
\emph{Ours}    & the process of playing shoot em' up videos games with friends                       \\ \bottomrule
\end{tabular}
\caption{Generated cases by our method that perfectly reproduce the target definitions. Note that we omit the ground-truth reference since they are exactly the same as the generated definitions.}
\end{table*}

\end{document}